\documentclass[letterpaper, 10 pt, conference]{ieeeconf}

\IEEEoverridecommandlockouts
\usepackage{cite}
\usepackage{amsmath,amssymb,amsfonts}

\usepackage[ruled,lined,linesnumbered]{algorithm2e}
\usepackage{algorithm2e}
\usepackage{algpseudocode}
\usepackage{comment}
\usepackage{tabularx}
\usepackage{graphicx}
\usepackage{textcomp}
\usepackage{xcolor}
\usepackage{float}
\usepackage{pifont}% http://ctan.org/pkg/pifont
\usepackage{comment}
\usepackage{xcolor}
\usepackage{url} 
\usepackage{balance}
\usepackage[some]{background}

\def\BibTeX{{\rm B\kern-.05em{\sc i\kern-.025em b}\kern-.08em
    T\kern-.1667em\lower.7ex\hbox{E}\kern-.125emX}}

\newcommand{\benchmark}{OROS}
\newcommand{\name}{REACT}

\begin{document}

\title{\LARGE \bf REACT: Multi Robot Energy-Aware Orchestrator for\\ Indoor Search and Rescue Critical Tasks}

\author{Fabio Maresca$^{*}{\dagger}$, Arnau Romero${\ddagger}$, Carmen Delgado${\ddagger}$,\\Vincenzo Sciancalepore$^{*}$, Josep Paradells${\dagger}$, Xavier Costa-P\'erez$^{*}{\ddagger}$
\thanks{$^{*}$F. Maresca, V. Sciancalepore, and X. Costa-P\'erez are with NEC Laboratories Europe GmbH
        {\tt\small \{name.surname\}@neclab.eu}}%
\thanks{$^{\ddagger}$A. Romero, C. Delgado, and X. Costa-P\'erez are with i2CAT Foundation and ICREA
        {\tt\small \{name.surname\}@i2cat.net}}%  
\thanks{$^{\dagger}$F. Maresca, and J. Paradells are with Universitat Politècnica de Catalunya.
        {\tt\small \{name.surname\}@upc.edu}}% 
}

\maketitle

\begin{abstract}
Smart factories enhance production efficiency and sustainability, but emergencies like human errors, machinery failures and natural disasters pose significant risks. In critical situations, such as fires or earthquakes, collaborative robots can assist first-responders by entering damaged buildings and locating missing persons, mitigating potential losses. 
Unlike previous solutions that overlook the critical aspect of energy management, in this paper we propose \textit{\name}, a smart energy-aware orchestrator that optimizes the exploration phase, ensuring prolonged operational time and effective area coverage. Our solution leverages a fleet of collaborative robots equipped with advanced sensors and communication capabilities to explore and navigate unknown indoor environments, such as smart factories affected by fires or earthquakes, with high density of obstacles. 
By leveraging real-time data exchange and cooperative algorithms, the robots dynamically adjust their paths, minimize redundant movements and reduce energy consumption. Extensive simulations confirm that our approach significantly improves the efficiency and reliability of search and rescue missions in complex indoor environments, improving the exploration rate by $10\%$ over existing methods and reaching a map coverage of $97\%$ under time critical operations, up to nearly $100\%$ under relaxed time constraint.
\end{abstract}

\section{Introduction}

The rise of smart factories, driven by the integration of technologies such as IoT, robotics, and Big Data processing with the aid of AI-based techniques in the Industry 4.0 has significantly improved efficiency, automation, and monitoring~\cite{hozdic2015smart}. However, these advancements do not eliminate the risk of emergencies, including machinery failures, fires, or external threats like earthquakes~\cite{tricomi2021resilient}. Ensuring worker safety in such scenarios requires effective search and rescue (SAR) operations.
Traditional SAR approaches rely on human intervention, often under perilous conditions, while shifting towards robotics integration would open up enhanced safety and efficiency benefits: collaborative connected robots can swiftly navigate and explore hazardous environments that would be otherwise inaccessible or too dangerous for human first responders~\cite{9220149}.

Unmanned SAR strategies have largely focused on outdoor environments, where the primary concern is often navigation over large, open areas with relatively sparse obstacles. These, however, fall short when applied to complex indoor environments characterized by high densities of obstacles and the need for precise navigation~\cite{wang2023development}. Recent research has emphasized the importance of real-time decision-making and obstacle mapping in such environments~\cite{10091098}, but such approaches often neglect a crucial aspect: optimal energy management. Energy consumption is a critical factor in SAR missions, as it directly influences the operational time and effectiveness of the robots~\cite{9779119,Delgado22}.
\begin{figure}[!t]
    \centering
    \includegraphics[width=0.95\linewidth]{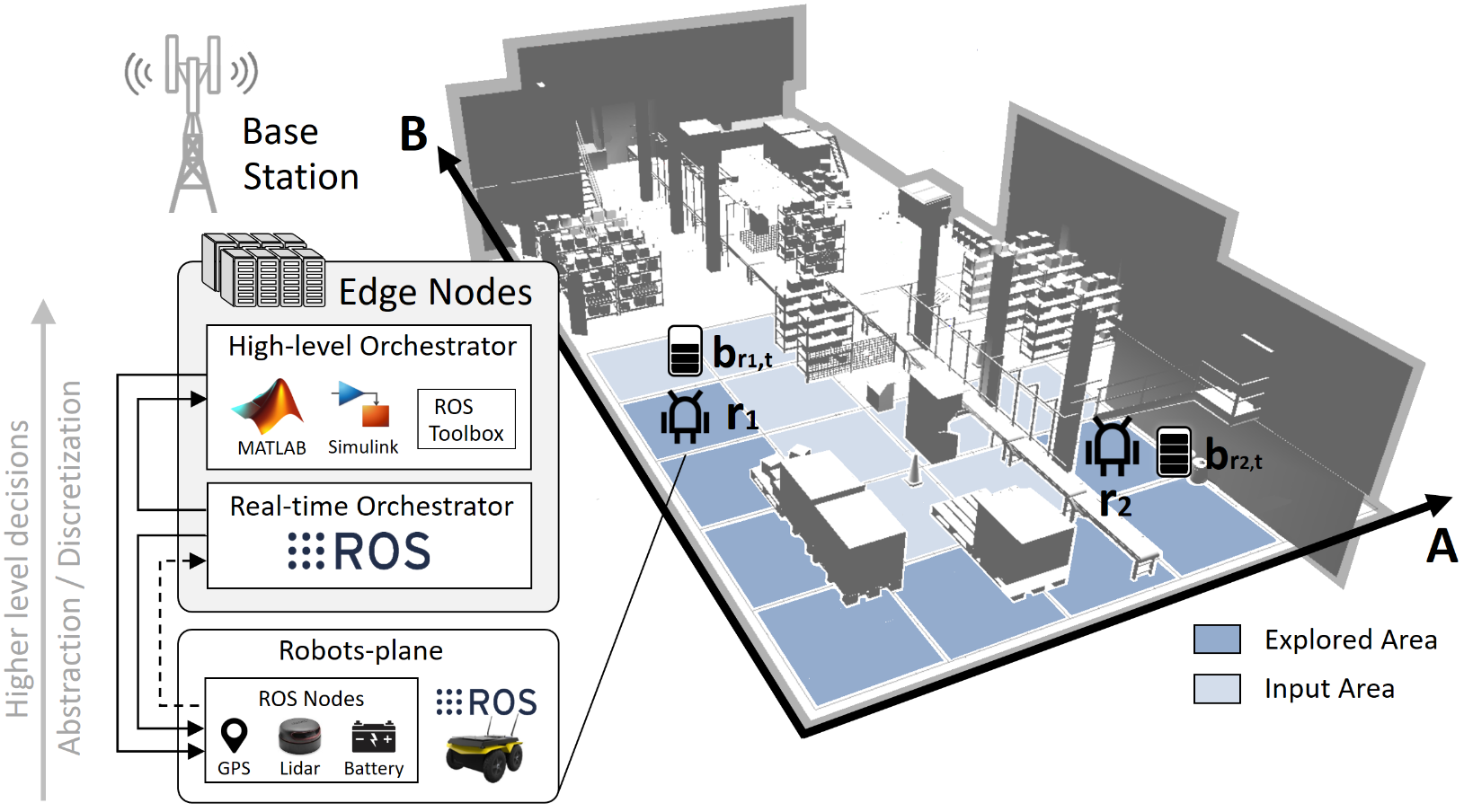}
    \caption{Smart Factory Multi Robot SAR scenario.}
    \label{fig:cover-image}
\end{figure}
Our paper addresses this gap by presenting a smart energy-aware indoor orchestrator that not only determines the optimal path planning for SAR operations in complex, obstacle-rich environments, like smart factories, warehouses or offices, but it also optimizes the robot resources. As shown in Fig. \ref{fig:cover-image}, our solution can directly trigger actions to guide robots in hard-to-explore scenarios.
Specifically, the proposed solution builds on top of our previous work \textit{OROS}\cite{OROS_TNSM}, an outdoors orchestration framework that minimizes mission-critical task completion times of 5G-connected robots by jointly optimizing robotic navigation and sensing together with infrastructure resources.

Particularly, depending on the density of obstacles, blind spots in the map and available energy on the robots, we flexibly adjust the individual path planning and corresponding resources. This work aims to extend the previous framework to operate in a new way, efficiently handling indoor scenarios that can be characterized by different challenges, such as the need for more reactive decisions and new logic for robot allocations, due to the presence of high-density obstacles. 
We present here a novel three-layer architecture ROS-compliant that is able to overcome state-of-the-art limitations.

\section{Related Works}

The adaptability and robustness of mobile robots have made them invaluable assets in hazardous environments~\cite{survey_SAR}. Consequently, they have been widely researched for SAR operations~\cite{RRT_SAR_2020}, which present distinct challenges depending on the scenario characteristics. Particularly, post-disaster indoor scenarios 
present a case where prior knowledge of the area might not be accurate, and with a high obstacle density due to rubble and debris. Coverage and response time are the most critical metrics to evaluate the operation performance \cite{SAR_metrics}. However, the coverage metric is closely related to the obstacle density in the scenario, as blind spots may arise where potential victims could be present. Partial or inaccurate prior knowledge of the exploration area, such as floor plans or sketches, might improve SAR operation performance~\cite{indoor_obstacles}. However, obstacle density has a high impact on both time and area coverage, since the robot has to slow down to avoid being jammed. Thus, the use of a real-time mission planner is still required to guarantee  reliability~\cite{10611179}. 

Several works also include energy performance as an additional key performance indicator, due to the limitations of battery capacity on autonomous mobile robots~\cite{batteries_amr}. Energy efficiency on robots can be enhanced by optimizing hardware usage (i.e. communications, sensors, computing processors), motion and navigation planning 
~\cite{IoRT_2020,hou_energy_2019,loganathan2023systematic}. Perception sensors and communications hardware frequency can be lowered or turned off whenever the robot does not require them. Meanwhile, traditional and most common motion and navigation planning tend to optimize energy efficiency by finding the shortest path in a flat terrain.  However, novel algorithms are proposed in unstructured terrains to minimize such energy consumption~\cite{zakharov2020energy}. Furthermore, energy efficiency can be enhanced by computation off-loading of high cost process required for the operation (e.g. victim detection through computer vision) using edge servers~\cite{qiu2022resource}.

In SAR scenarios, the mission outcome depends on various characteristics, ranging from the environment configuration to the number and configuration of the robots. Thus, the use of ROS middleware and the Gazebo\footnote{More information is avaiable at: \url{https://gazebosim.org/}} simulator are well-established tools for evaluating robot performance in SAR use cases~\cite{coop_robot_ROS_2020, mirzae2020ros}. A system built using ROS consists of several processes, named ROS nodes, connected at runtime in a peer-to-peer topology. Furthermore, the software allows the simulation of multiple robot systems operation, which are used to fasten response time and increase area coverage~\cite{multi_robot_2021}. 

Robot coordination in multi-robot systems is necessary to avoid redundancies and avoid collisions. Upon evaluating the overall mission performance, task planning can be computed optimally using a centralized cloud or edge-based coordinator~\cite{cloud_multiple}. In fact, the overall SAR operation efficiency can be enhanced by using a centralized orchestration scheme for robot fleet path planning, leveraging edge computing and
communication technologies. 
The authors in~\cite{L_NORM_2023} use Wi-Fi, however, first responders may not have access to already deployed technologies and would benefit from deploying their own cellular network.
In fact, the authors in~\cite{OROS_TNSM} propose a joint 5G and robot orchestration logic that indicates optimal path planning of a robot fleet, as well as, robot hardware usage (i.e. sensors, communication peripherals) and battery charging priorities using a charging station. However, obstacles where considered in high level task planning when they did not allow passage from one area to another, assuming no blind spots were formed. However, this approach operates with a simplified, highly discretized view of the SAR operation, while scenarios with numerous objects require a more detailed and realistic view. This necessitates a higher frequency of decision-making based on each robot's assigned task.
A multi-layered system can be used to enhance high level decision-making algorithms such as a robot orchestrator. In such architectures, robots provide exploration data (e.g. costmaps) that a group of middle-layers merge and use to improve target exploration determination~\cite{CURE_2024}.

In this work, we use the lessons learnt from \cite{OROS_TNSM} and \cite{CURE_2024} and build our own energy-aware three-layer multi-robot indoor orchestrator, which, to the best of our knowledge, is the only one that guarantees and energy-efficient path exploration and resource management.

 \section{\name: Concept, System and Architecture}

As depicted in Fig.~\ref{fig:cover-image}, we consider a post-disaster indoor scenario, such as smart factories, where a fleet of robots is deployed for SAR operations. In these scenarios, while the overall size of the exploration area may be known based on building dimensions, the interior conditions are unknown. The robots collaborate using 5G cellular connectivity, which can be provided either by the smart factory's infrastructure or by a dedicated base station. In these indoor SAR scenarios, a higher density of obstacles is expected compared to outdoor operations. These obstacles may consist of debris, rubble, or factory machinery, obstructing the robots' line of sight and creating blind spots where disaster victims could be trapped. Consequently, thorough and detailed area exploration is required to ensure no critical zones are overlooked.

\begin{figure}[t!]
\includegraphics[width=0.5\textwidth]{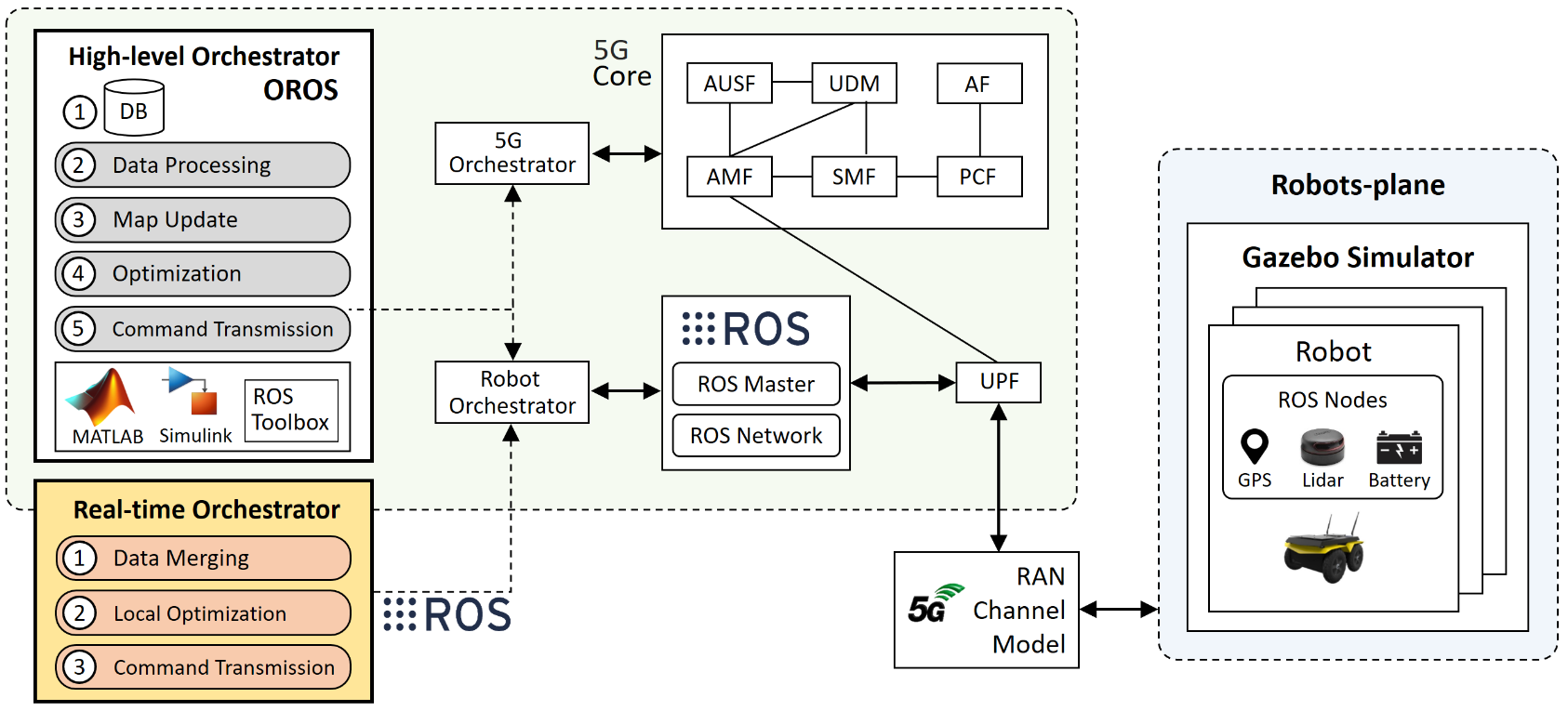}
\caption{\textit{\name} architecture: components, technologies used, interfaces and communication links.}
\label{fig:architecture}
\end{figure}

\textit{\name} is a three-layered architecture integrating the \textit{\benchmark} plane and the robot plane, with a middle layer handling dynamic decision-making and real-time instructions.
Thanks to this added layer and their corresponding interactions, \textit{\name} is able to guarantee reliability and enhanced performance, both in terms of coverage and energy savings. While the High-level Orchestrator layer and the Real-time Orchestrator layer run at the edge, the Robots-plane layer is directly deployed on the robots, thanks to ROS. Fig.~\ref{fig:architecture} shows the \textit{\name } architecture, that employs ROS for the Robots-plane and the Real-time planner, and MATLAB and Simulink for the logic of \textit{OROS}. The ROS Toolboxes allow interfacing between the components. Connectivity across the system is achieved through a simulated 5G channel link, enabling IP-based communication~\cite{TR38.900}.
In the following, we describe each layer in detail.

\subsection{High-level Orchestrator}
The High-level Orchestrator, also known as \textit{OROS}, is in charge of jointly optimizing robotic navigation and sensing together
with infrastructure resources. For this purpose, this orchestrator establishes a grid to monitor visited areas and plan the optimal route, while also managing the robot resources. In terms of the path planning, it directs the fleet toward high-level target points, which are the centroids of the next areas to reach, and computes energy-saving policies, such as instructions to turn off specific sensors when revisiting known areas.

Specifically, it discretizes the 2D surface into a grid $\mathcal{G} =\{g_{a,b}, \forall (a,b) \in (A,B) \}$,  where each element $g_{a,b}$ needs to be explored by at least one robot. To keep track of the multi-robot exploration, \textit{OROS} introduces $e_{t,a,b}$ as a binary variable, indicating whether  the area unit $g_{a,b}$ has been already explored at time $t \in \mathcal{T}$. In addition, \textit{OROS} considers the  battery level variable $b_{r,t}$ for each robot at time $t \in \mathcal{T}$ to solve the optimization problem defined as:  

\noindent \textbf{Problem}~\texttt{OROS ($ \mathcal{T}$)} :
\begin{flalign}
  \quad\quad & \text{max} 	\sum_{t \in \mathcal{T}} \sum_{(a,b) \in (A,B)} e_{t,a,b}  + \sigma \sum_{r \in \mathcal{R}}  b_{r,|\mathcal{T}|},
  \label{prob:OROS}
\end{flalign}
where $\sigma$ is a normalization constant. The output of the optimization problem is a high-level task plan that specifies a target subarea  $g_{a,b} \in \mathcal{G}$ to navigate to at a given speed for each robot $r \in \mathcal{R}$ at each time $t \in \mathcal{T}$. If the target subarea is unexplored, the \textit{OROS} optimization algorithm instructs robots to use their perception sensors, processors, and communication resources. Otherwise it directs to turn them off. The collection of raw data, the solution of \textit{Problem} \ref{prob:OROS}, and the resulting instructions are synthesized in the instruction \texttt{Call\_OROS} in Alg.~\ref{alg:alg1}.

\subsection{Real-time Orchestrator}

The Real-time Orchestrator is the middle-layer that operates with higher data resolution and performs real-time multi-robot path planning, collision avoidance and robot local allocation for new temporary targets during task performance. Specifically, it implements the steps described in Alg.~\ref{alg:alg1}. For the duration of the SAR mission, it integrates environment data from the robot fleet into a merged global occupancy grid. 
Analysing this costmap, the unknown regions within each robot's area are identified and their centroids are defined as new local targets after assessing their accessibility (line~5).

Once unknown regions are identified, the planner allocates the robots to local targets (Alg.~\ref{alg:alg1}, line 6), choosing the one with better positioning and battery level when multiple are concurring for the same zone. 

Areas of interest are marked with a priority based on their size and predicted contribution to the path planning update. Priority is higher when a target is located near neighboring unexplored areas, as these regions may open access to new paths that were not previously planned. By prioritizing these targets, the system can facilitate the discovery of alternative routes, potentially enhancing overall mission efficiency and adaptability. 
Non-priority blind spots are postponed to favour a greedy exploration strategy and to achieve rapid coverage. 
After achieving all local targets, the Real-time orchestrator updates the percentage of explored area $c_{a,b}$ to the \textit{OROS} orchestrator, that is then marked as completely explored when reaching a certain threshold $\delta$.

Finally, when all the elements of the grid $\mathcal{G}$ are completely explored, the attention is redirected to those secondary targets left behind. These blind spots are often disregarded due to their low probability of containing a person or their difficult accessibility. Despite this, robots are tasked with attempting to reach these areas, even if navigating through them requires more time and poses a risk of getting stuck (the reason why \textit{\name} prefers to do it at the end).
The trigger event is formalized in Equation \ref{eq:continuation_condition}:
\begin{equation}
\sum_{a \in A} \sum_{b \in B} e_{t,a,b} = |A B| \hspace{0.5cm} t \in \mathcal{T} 
\label{eq:continuation_condition}
\end{equation}

\begin{algorithm}[!t]
\small
\SetKwInOut{Input}{Input}
\SetKwInOut{Output}{Output}
\SetKwInOut{Return}{return}
\SetKwInOut{Initialize}{Initialize}
\SetKwInOut{Procedure}{Procedure}
\SetKwInOut{Goto}{Go To}
\Input{$ \mathcal{T}, \mathcal{R}, \mathcal{G}, P_{robot};$}
\Initialize{ $e_{0,a,b}, b_{r,0} ; $}
\Procedure{}
\While{ $t < |\mathcal{T}|$ }{
    \textbf{For each} {$r \in \mathcal{R}$:} \\
    \Indp
    \While{ $c_{a,b} < \delta$ }{ 
        UPDATE merged\_map\;
        UPDATE local\_targets\;
        Allocate\_robot\_to\_target()\;
        MOVE Robot\;
    }
    UPDATE $e_{a,b}$\;
    \Indm
    \texttt{Call\_OROS}\;
    \If{Equation \ref{eq:continuation_condition} holds,}{
        \If{$b_{r,t+t_{dpi}}>0$}{
            Deferred\_priority\_inspection()\;
        }
    }
}
\Output{ $coverage, path\_planning;$}
\caption{\name}
\label{alg:alg1}
\end{algorithm}

The planner checks whether the operation time limit and the robots' battery are sufficient to visit the new final targets.
Thus, the Real-time orchestrator estimates the needed time for the deferred priority inspection $t_{dpi}$ and the battery level upon attainment, $b_{r,t+t_{dpi}}$. 
If these parameters meet the conditions for continuing the mission, it commands the fleet to reach the final targets, allocating robots based on their positions and battery levels (Alg.~\ref{alg:alg1}, line 13).

This approach prevents early slowdowns introduced by a Naïve meticulous exploration and allows the coverage to progress efficiently while minimizing the time lost in challenging areas.
As shown in the results for the Naïve counterpart, a robot may struggle to return to the previous plan when it enters narrow burrows, slowing exploration and inefficiently depleting the battery. This risks failing to access subsequent areas of the grid and covering the entire map within operational limits. Consequently, less efficient exploration points are deferred only if resources are exceeded after greedily covering most of the map. 
Additionally, this approach relaxes the previous assumption of a fixed map size known a priori. Instead of requiring a predetermined perimeter, we now aim to explore environments of indefinite extent, accommodating changes in topology due to emergencies.

\begin{figure}[t!]
    \centering
    \includegraphics[width=1\linewidth]{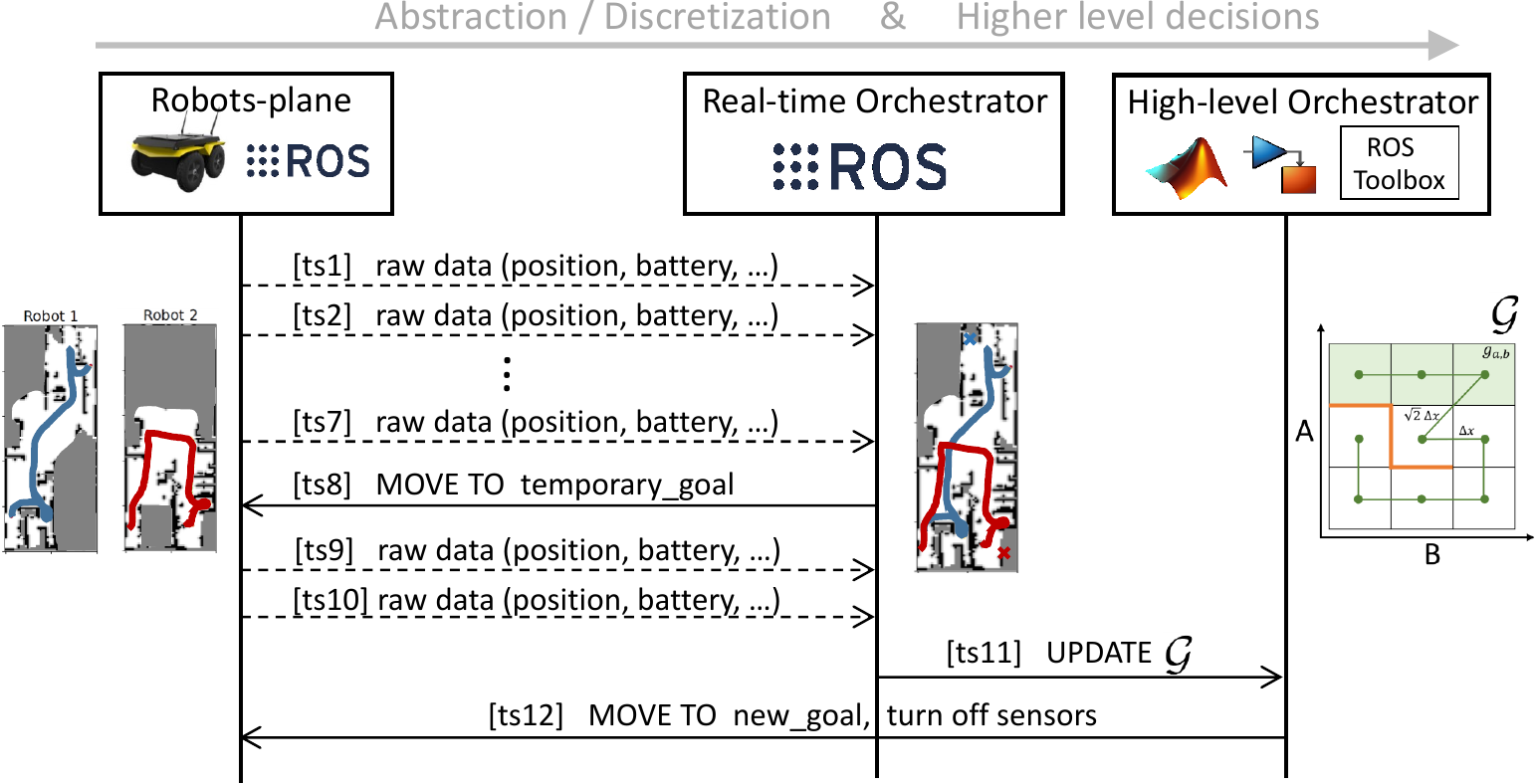}
    \caption{The sequence diagram highlights the exchange of messages among the three layers within a time frame reserved to explore the current area. The frequency of occurrence is emulated with example timestamps \textit{ts}.}
    \label{fig:sequence_diagram}
\end{figure}

\subsection{Robots-plane}

Finally, at the bottom layer, we have a predefined set of autonomous mobile robots equipped with the necessary hardware to navigate and explore the environment using perception sensors (e.g., LiDAR, cameras), processors, and communication capabilities. These robots are simulated in the Gazebo environment with ROS support, where they share information through messages in the form of ROS topics. These include messages related to sensor data, control commands, and robot status updates. Each robot continuously collects LiDAR points to map objects that either hinder transitions between areas or partially obstruct the robot’s visibility of the current area.

\subsection{Interaction among layers}

The interaction among the three layers within a time frame reserved to explore the current area (e.g., 10s) are illustrated in the sequence diagram in Fig.~\ref{fig:sequence_diagram}. The information exchange between robots and the middle-layer planner is bidirectional: while the firsts continuously transmit raw data at each timestamp, the Real-time orchestrator provides updated directives when an obstacle that obstructs visibility is detected. The third layer is engaged at the end of every \texttt{Call\_OROS} (Alg.~\ref{alg:alg1}, line 10) to plan a new route if necessary and to issue commands to the robots, including instructions for new target areas and operational modes (e.g., power-saving mode with sensors off).
The sequence diagram primarily depicts the message flows, while the internal logic of each component is represented through their respective views (e.g., individual costmaps for the robots, a merged costmap for the middle layer, and a grid for the high-level layer).
Raw data transmission is depicted with dashed lines, while instructions to perform are defined as straight lines in both Fig.~\ref{fig:cover-image} and Fig.~\ref{fig:sequence_diagram}.

\section{Performance Evaluation}

Hereafter, we evaluate the performance of \textit{\name} when multiple robots are deployed in an unknown indoor scenario. We first introduce the scenario setup, and then we evaluate the performance of the proposed solution by evaluating both the coverage and the energy savings obtained in relation to the operation time expressed in seconds. The experiments are first conducted with a single robot to analyze the algorithm, and later in a multi-robot scenario to showcase the full improvement obtained.

\subsection{Scenario setup}
\label{sec:setup}

The evaluation of \textit{\name} was conducted through extensive simulations in Gazebo, designed on a virtual machine running ROS, Gazebo, Rviz and all the necessary libraries for robot simulation and visualization. 
The baseline \textit{\benchmark} logic was computed on a second virtual machine representing an edge node, hosting MATLAB and Simulink services, as well as ROS-bridging components to exchange messages with the robots.
The Real-time orchestrator layer was implemented via \textit{rosnodes} on the edge, publishing \textit{rostopics} for both the robots and the Simulink-assisted node, as earlier explained.

The evaluation was performed in a realistic industrial environment: an extended 20x40-meter factory floor obtained from an open-source dataset\footnote{https://github.com/mlherd/Dataset-of-Gazebo-Worlds-Models-and-Maps/tree/master/worlds/factory}, which contains various obstacles, including two conveyor belts, shelves, and command stations, which present significant navigation challenges, and where we have included additional blind spots and narrow tunnels, making it of approximately 20x50meters map.
The robots' starting points are placed in the tunnel at the beginning of the map.
Specifically, we use Jackal wheeled robots\footnote{https://clearpathrobotics.com/jackal-small-unmanned-ground-vehicle/}, each equipped with an 8-meter radius LiDAR and a battery with an energy capacity of 8245.96 J, whose energy profile is detailed in \cite{10611179}.
We compare \textit{\name} with the \textit{\benchmark} solution, which does not include the Real-time orchestrator layer, but has an energy-awareness component, since other state-of-the-art works do not include this energy plane. In fact, to show the energy performance, we compare with the case where sensors, processors and communications cannot be turned off.

\subsection{Single-Robot Evaluation}

\begin{figure}[t!]
        \includegraphics[width=\linewidth]{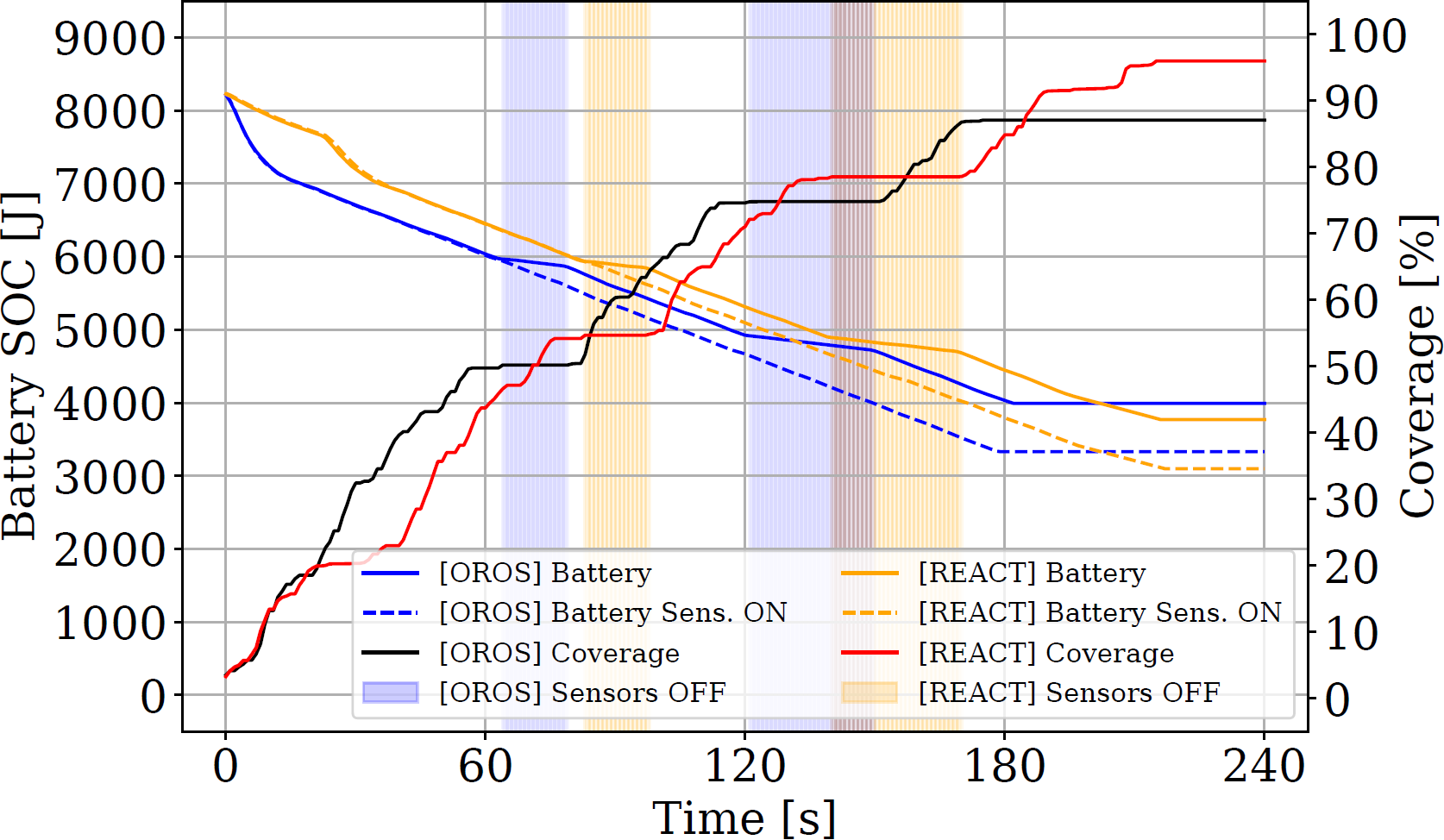}   
        \caption{Battery SOC and Coverage for 1 robot using \textit{\benchmark}  and  \textit{\name}. The maximum Operation time is T=240. \textit{"Battery Sens. ON"} refers to a non-optimized management of the sensors, keeping them always ON.}
        \label{fig:comparison_exploration_battery_1}
\end{figure}

In Fig.~\ref{fig:comparison_exploration_battery_1} we show the battery State of Charge (SOC) and coverage (in percentage) for a single robot in a single experiment (since it is deemed unnecessary to average the battery level) in the factory presented in Sec. \ref{sec:setup}. We evaluate the two approaches, namely \textit{\benchmark} and \textit{\name}, both with and without the ability to turn off sensors when not needed (i.e., when visiting already explored areas) to save energy. In the \textit{Always ON} mode, all sensors and processors remain active throughout the experiment. As shown in the figure, \textit{\benchmark} guarantees $\sim$$88.5\%$ exploration, accessing all the areas defined on the map in 170s disregarding blind spots, while \textit{\name} explores further, reaching a total of $97\%$ coverage around 15s later and then finishes exploring at second 213.
Furthermore, for the energy-aware strategies, the comparison between straight lines and the corresponding dashed ones shows the amount of saved energy while optimizing the usage of the sensors. These are, respectively, the battery trends when turning off the sensors  and keeping them always ON in known areas, that is the case of the time intervals $\sim$65-75s and $\sim$120-150s for \textit{\benchmark}.

\begin{figure}[t!]
    \centering
    \includegraphics[width=0.9\linewidth]{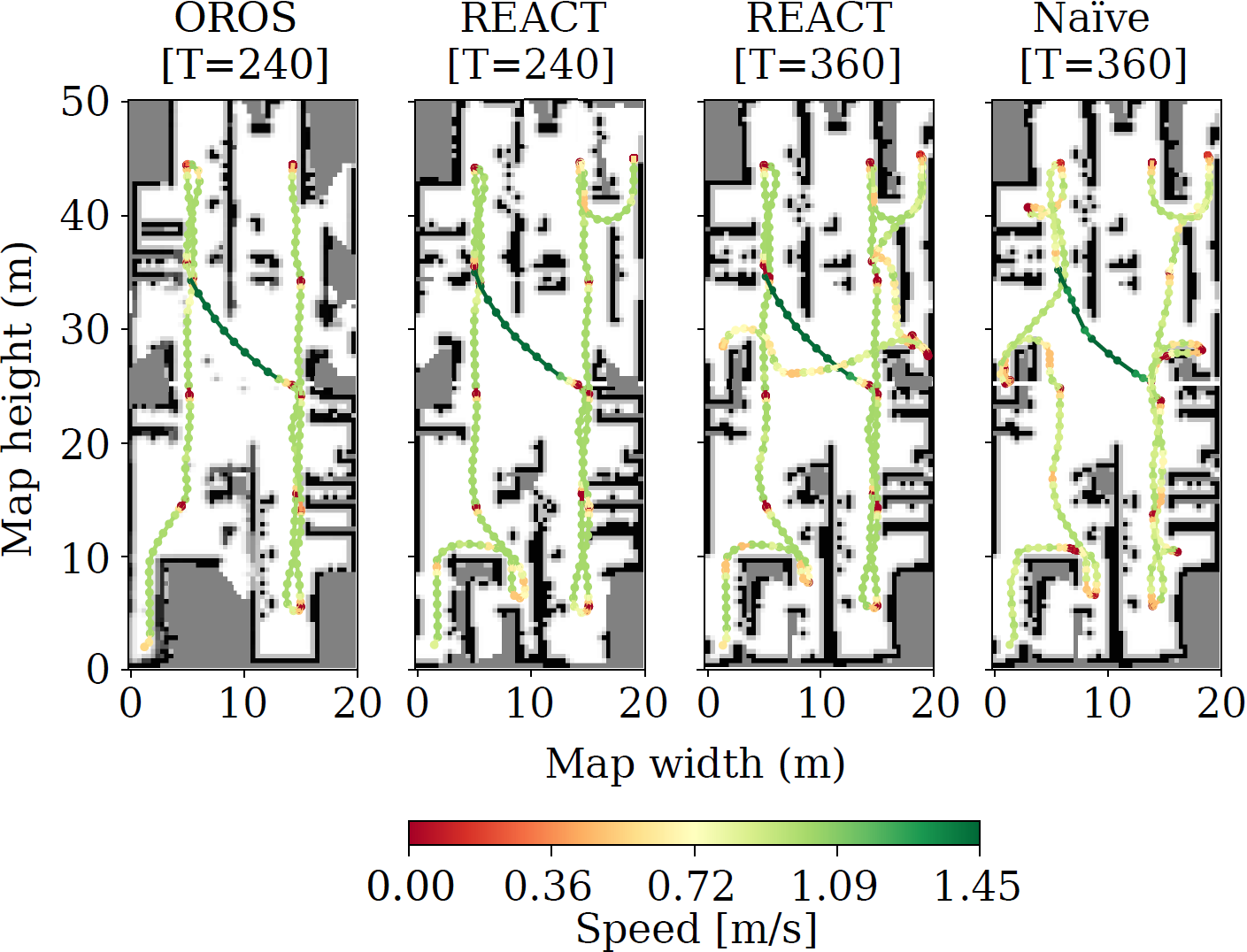}
    \caption{Trajectory performed by 1 robot, deploying different methods.}
    \label{fig:trajectories_1r}
\end{figure}

The trajectories followed by the robots are depicted in Fig.~\ref{fig:trajectories_1r}. First, it can be seen that the improvement of \textit{\name} is mainly reached by accessing the last area (upper-right). Second, the energy-savings due to the energy-awareness early explained, can easily be understood by looking at the paths followed by the robots in the areas centered in (5,35) and (5,45) and (15,5) and (15,15) (which corresponds to the $\sim$65-75s and $\sim$120-150s for \textit{\benchmark}). In these areas, robots go through already explored areas and the sensors are turned off, reason why the exploration does not improve and the battery discharges slower than the respective \textit{Always ON} counterpart, saving a total of $\sim$$750 J$ at the end of the simulation. Few seconds later the same happens for \textit{\name}, that visits the same areas in the same order but with a delay of some seconds needed to explore the blind spots in the first area (lower left in the map in Fig.~\ref{fig:trajectories_1r}). 
Thus, setting the maximum operation time to $T=240s$ (assuming the mission is highly critical), both \textit{\benchmark} and \textit{\name} can access all the areas of the grid deploying a single robot. However, the time is not enough to explore secondary blind spots left behind, which are reachable extending $T$ to 360s, as shown in Fig.~\ref{fig:comparison_exploration_battery_1}. 

\begin{figure}[t!]
        \centering
        \includegraphics[width=0.97\linewidth]{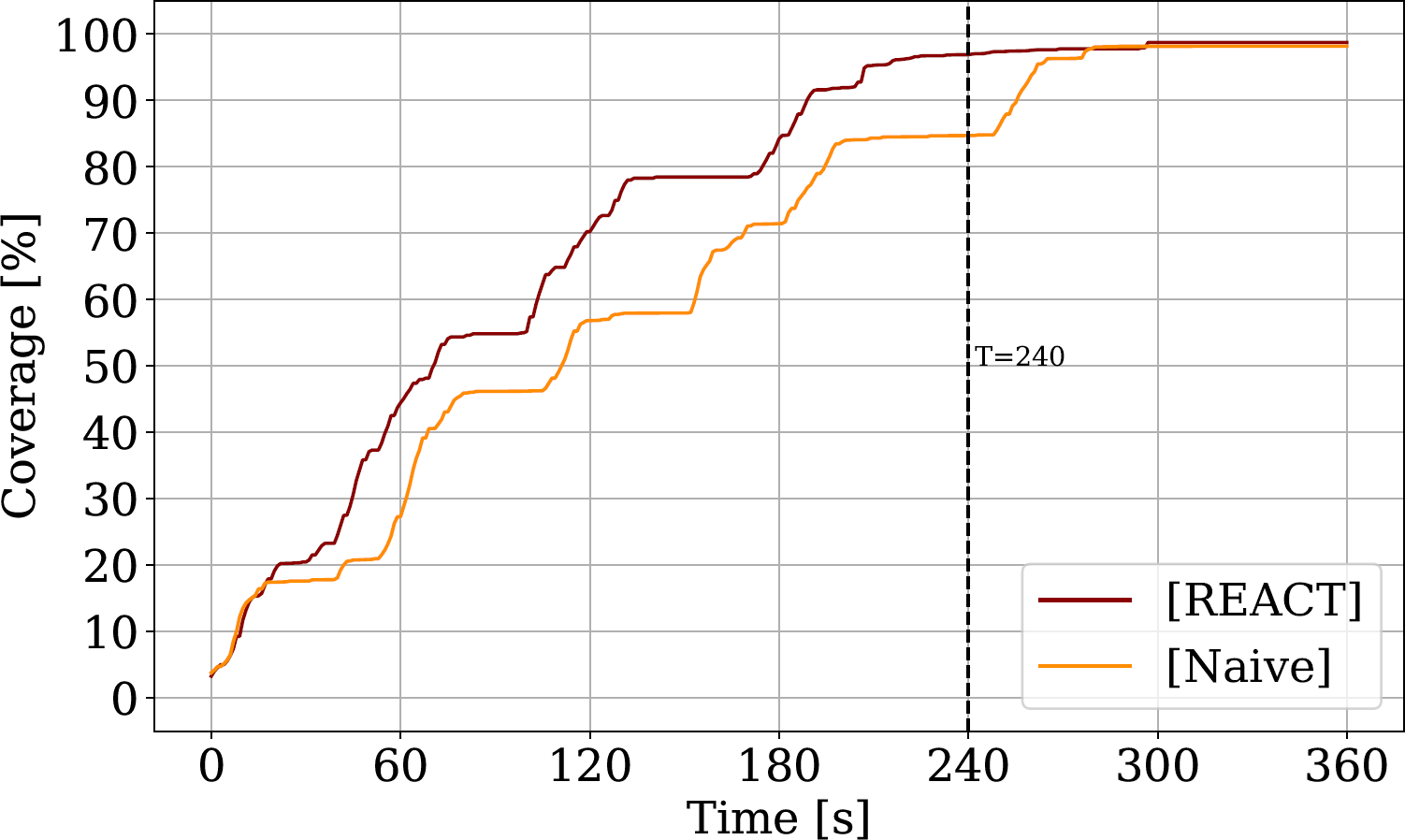}
        \caption{Coverage for 1 robot showing differences in high-critical operations performance between \textit{\name} and its Naïve counterpart.}
        \label{fig:comparison_exploration_battery_2}
\end{figure}

In fact, to illustrate the optimal decisions of \textit{\name}, in this plot we also show a Naïve solution that does not prioritize the overall map coverage, treating all obstacles indistinctly. The latter is characterized by a slower coverage rate, as depicted in Fig.~\ref{fig:comparison_exploration_battery_2}, which may result in failure to explore all the areas defined on the map when the mission duration is critical, such as $T=240s$.

\subsection{Multi-Robot Evaluation}

\begin{table*}[ht]
\centering
 \caption{Energy consumption per robot and relative savings when setting $\mathcal{R}=\{1,2,3\}$}
\label{tab:energy_consumption}
\begin{tabular}{|l||c|c|c||c|c|c|}
\hline
 &
  \textbf{\textit{\benchmark}} &
  \textbf{\textit{\benchmark} Always ON} &
  \textbf{\textit{\benchmark} Savings} &
  \textbf{\textit{\name}} &
  \textbf{\textit{\name} Always ON} &
  \textbf{\textit{\name} Savings} \\ \hline
$\mathcal{R}$=1 &
  4238.01 J ($r_1$)&
  4914.60 J ($r_1$)&
  676.59 J (14.2\%) ($r_1$)&
  4398.37 J ($r_1$)&
  5147.92 J ($r_1$)&
  749.55 J (14.5\%) ($r_1$)\\ \hline
$\mathcal{R}$=2 &
  \begin{tabular}[c]{@{}c@{}}3516.80 J ($r_1$)\\ 3229.89 J ($r_2$)\end{tabular} &
  \begin{tabular}[c]{@{}c@{}}3516.80 J ($r_1$)\\ 3900.06 J ($r_2$)\end{tabular} &
  \begin{tabular}[c]{@{}c@{}}0 J ($r_1$) \\ 670.17 J (17.2\%) ($r_2$)\end{tabular} &
  \begin{tabular}[c]{@{}c@{}}3499.54 J ($r_1$)\\ 3386.22 J ($r_2$)\end{tabular} &
  \begin{tabular}[c]{@{}c@{}}3499.54 J ($r_1$)\\ 4301.65 J ($r_2$)\end{tabular} &
  \begin{tabular}[c]{@{}c@{}}0 J ($r_1$)\\ 915.43 J (21.3\%) ($r_2$)\end{tabular} \\ \hline
$\mathcal{R}$=3 &
  \begin{tabular}[c]{@{}c@{}}3028.65 J ($r_1$)\\2799.20 J ($r_2$)\\2766.96 J ($r_3$)\end{tabular} &
  \begin{tabular}[c]{@{}c@{}}3028.65 J ($r_1$)\\2799.20 J ($r_2$)\\2766.96 J ($r_3$)\end{tabular} &
  0 J &
  \begin{tabular}[c]{@{}c@{}}2986.10 J ($r_1$)\\3093.18 J ($r_2$)\\2772.64 J ($r_3$)\end{tabular} &
  \begin{tabular}[c]{@{}c@{}}2986.10 J ($r_1$)\\3093.18 J ($r_2$)\\2772.64 J ($r_3$)\end{tabular} &
   0 J \\ \hline
\end{tabular}
\end{table*}

In order to evaluate the performance of multi-robot deployments, we have deployed different sets of robots. Performance, particularly in terms of operational time, improves as more robots are deployed, with the peak efficiency observed when utilizing three agents. Introducing a fourth robot can result in degraded performance due to an increased competition for targets and overlapping trajectories in shared areas. Indeed, placing the robots within the same initial area, forced them to share some initial paths.
Thus, in the factory scenario under study, three robots represent the maximum amount that still improves the performance.

\begin{figure}[t!]
    \centering
    \includegraphics[width=0.80\linewidth]{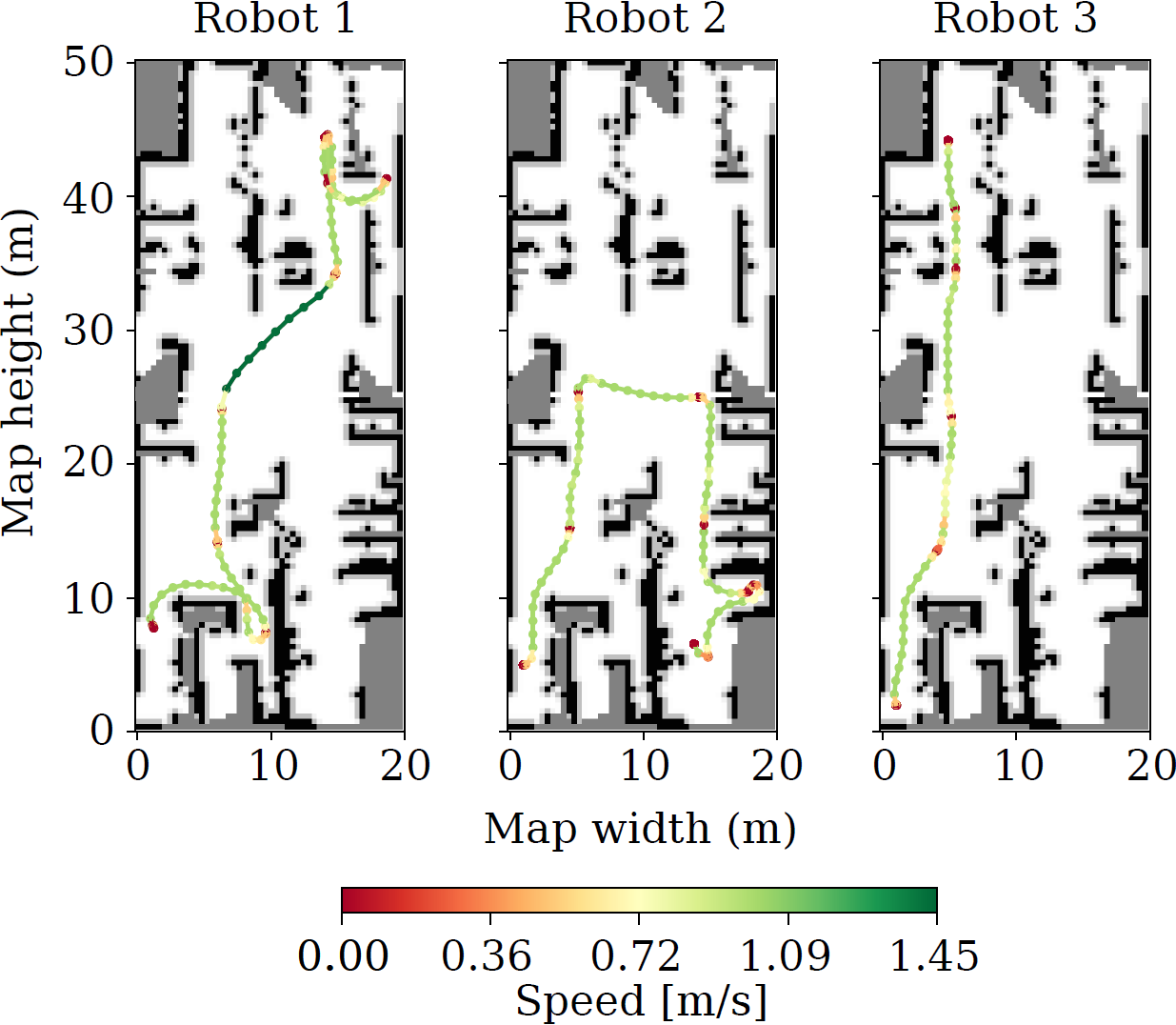}
    \caption{Three robot trajectories planned for \textit{\name}.}
    \label{fig:3robots_positions}
\end{figure}

\begin{figure}[t!]
    \centering
    \includegraphics[width=0.9\linewidth]{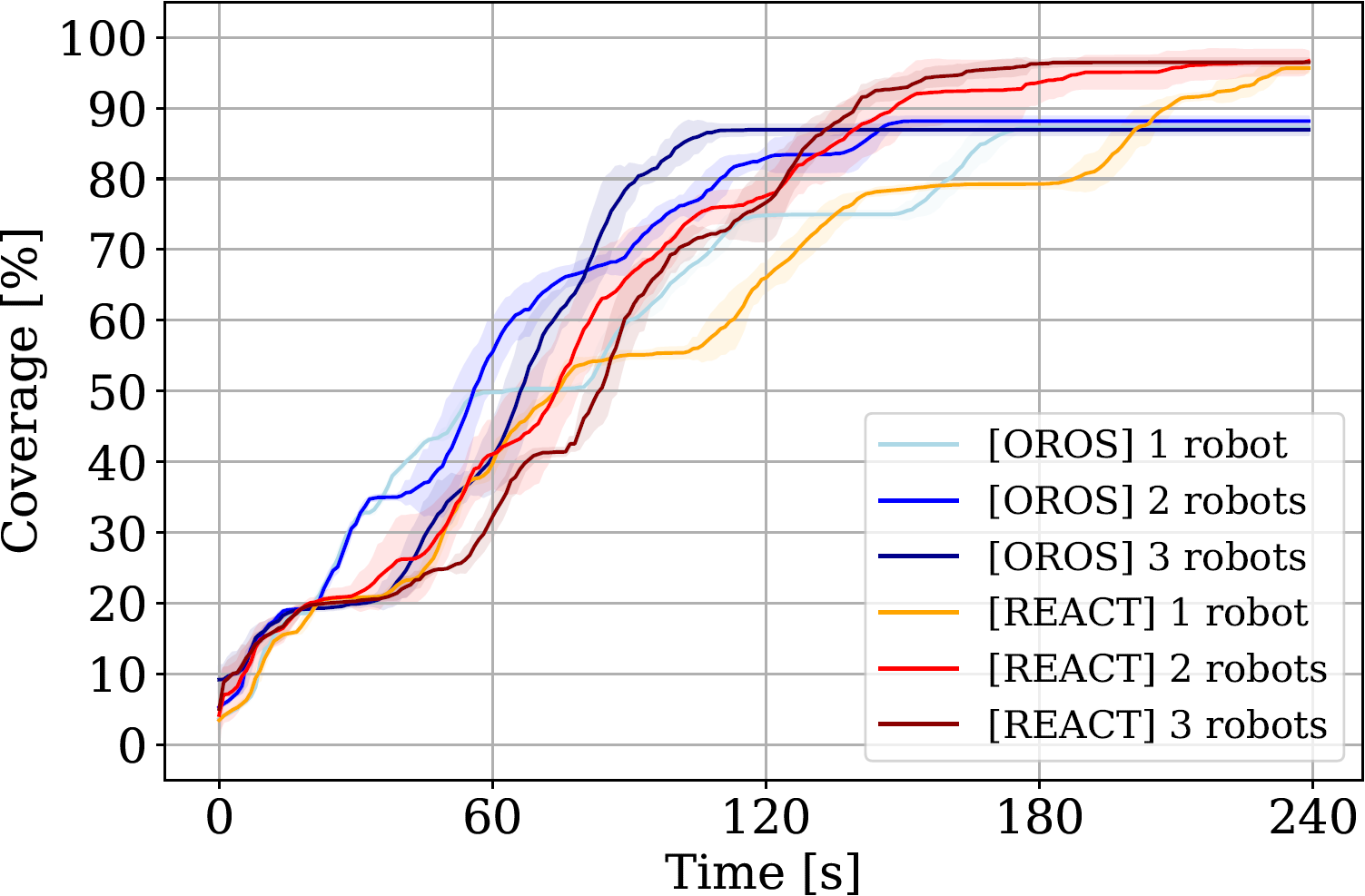}
    \caption{Coverage achieved per time by using multi-robots. The results compare the deployment of $\mathcal{R}=\{1,2,3\}$ robots.}
    \label{fig:explorations3robots}
\end{figure}

Fig.~\ref{fig:3robots_positions} illustrates the trajectories followed by \textit{\name}, which is able to efficiently assign goal positions such that the robots avoid revisiting previously explored areas, so that energy is not wasted. In order to better understand the behaviour, in Fig.~\ref{fig:explorations3robots} we show the average exploration over time achieved by the \textit{\benchmark} and the \textit{\name} strategies when varying the number of robots. The experiments are characterized by small variations in exploration times, due to the variation in the robot's speed and waiting times in proximity to obstacles. Given this component of randomness, albeit minimal, each experiment is repeated 5 times and averaged. Therefore, the mean (line) and standard deviation (colored area) are reported.

As can be seen, for \textit{\benchmark}, the time required for three robots to cover its maximum exploitable area is 110s, compared to 150s and 175s for two and one robot, respectively. The number of robots does not affect the overall exploration coverage, which remains consistent at approximately $88\%$ under this approach. However, \textit{\name} is able to reach up to $97\%$ coverage, by employing 160s with three robots, while two and one robot required 190s and 210s, respectively.
The higher standard deviation observed in this case is attributable to the robots’ interaction with obstacles. The robots experience significant deceleration when navigating in close proximity to obstacles, and this algorithm prompts them to attempt traversal through narrow paths. Consequently, these subtle differences in distance from objects result in slightly differing outcomes in each trial.

In order to show the energy savings obtained by \textit{\name}, Table \ref{tab:energy_consumption} provides comprehensive details on battery consumption. Using an energy-aware approach is always crucial, as can be seen by both savings. However, not only \textit{\name} is able to explore more area, but also save more energy.
Let us note that in the case of three robots, as shown in Fig.~\ref{fig:3robots_positions}, every robot always move through new areas, and therefore, the framework does not allow to disable any robot sensors, reporting no energy benefits (0J saved). In the case of using 2 robots, while 1 robot tries to move through always new areas, the other has to move through already explored ones, but being able to save some energy (up to $21\%$).

\section{Conclusions and Future Work}
\label{sec:conclusion}

In this paper, we presented \textit{\name}, a novel smart energy-aware orchestrator for collaborative connected robots, designed to optimize SAR missions in complex indoor environments such as smart factories and other obstacle-dense settings. Our approach not only addresses the limitations of previous outdoor-focused solutions but also provides a tailored energy management strategy that extends the operational time of robotic fleets, prioritizing specific areas. 
The orchestrator enables the robots to explore large, hazardous areas while minimizing redundant movements and energy consumption through dynamic path planning and real-time cooperation. Our simulations, deploying a fleet of wheeled Jackal robots in Gazebo,  demonstrated improvements in the exploration rate of approximately $10\%$ compared to state-of-the-art methods, showcasing its effectiveness in scenarios where timely and energy-efficient exploration is critical for successful SAR missions. The proposed solution also illustrates how advancements in real-time data exchange and cooperative algorithms can be leveraged to create adaptive strategies that respond to evolving environmental conditions and energy and time constraints.

Future research will integrate machine learning for autonomous decision-making, conducting real-world testing in smart factories, and exploring hybrid energy sources like solar or wireless power to extend robot missions. 

\section*{Acknowledgment}
This work was partially supported by the SNS JU under the European Union's Horizon Europe research and innovation programme under Grant Agreement No. 101192521 (MultiX), by EU FP for Research and Innovation Horizon 2020 under Grant Agreements No. 956670 (5GSmartFact), by the Spanish Ministry of Economic Affairs and Digital Transformation and the European Union – NextGeneration EU (Call UNICO I+D 5G 2021, ref. number TSI-063000-2021-6 and ref. number TSI-063000-2021-14), and by the CERCA Programme from the Generalitat de Catalunya.

\balance
\bibliographystyle{IEEEtran}
\bibliography{biblio}

\end{document}